\documentclass{article}

\usepackage{microtype}
\usepackage{graphicx}
\usepackage{subfigure}
\usepackage{booktabs} 

\usepackage{geometry}

\usepackage{fancyref}
\usepackage{fancyhdr}
\usepackage{amsmath}
\graphicspath{{assets/}} 
\DeclareMathAlphabet\mathcalbf{OMS}{cmsy}{b}{n}
\usepackage{mathabx}
\usepackage{mathtools}
\usepackage{cancel}
\usepackage{caption}
\usepackage{floatrow}
\usepackage[T1]{fontenc}
\usepackage{multirow}
\usepackage{bbm}
\usepackage[hang,flushmargin]{footmisc}
\usepackage{appendix}

\def\eqref#1{equation~\ref{#1}}

\def\1{\bm{1}}

\DeclareMathAlphabet{\mathsfit}{\encodingdefault}{\sfdefault}{m}{sl}
\SetMathAlphabet{\mathsfit}{bold}{\encodingdefault}{\sfdefault}{bx}{n}

\DeclareMathOperator*{\argmin}{arg\,min}

\usepackage{hyperref}

\usepackage[accepted]{icml2021}

\icmltitlerunning{Dynamic Value Estimation for Single-Task
            Multi-Scene Reinforcement Learning}

\begin{document}

\twocolumn[
\icmltitle{Sparse Attention Guided Dynamic Value Estimation for Single-Task
            Multi-Scene Reinforcement Learning}




\begin{icmlauthorlist}
\icmlauthor{Jaskirat Singh}{anu}
\icmlauthor{Liang Zheng}{anu}
\end{icmlauthorlist}

\icmlaffiliation{anu}{Research School of Computer Science, Australian National University, Canberra, Australia }

\icmlcorrespondingauthor{Jaskirat Singh}{jaskirat.singh@anu.edu.au}

\icmlkeywords{Multi-Scene Reinforcement Learning, Variance Reduction, Dynamic Value Estimation}

\vskip 0.3in
]



\printAffiliationsAndNotice{}  

\begin{abstract}
Training deep reinforcement learning agents on environments with multiple levels / scenes from the \emph{same task}, has become essential for many applications aiming to achieve generalization and domain transfer from simulation to the real world \cite{wortsman2019learning,cobbe2019quantifying}. While such a strategy is helpful with generalization, the use of multiple scenes significantly increases the variance of samples collected for policy gradient computations. Current methods, effectively continue to view this collection of scenes as a single Markov decision process (MDP) and thus, learn a scene-generic value function $V(s)$. However, we show that the sample variance for a multi-scene environment is best minimized by treating each scene as a distinct MDP, and then learning a joint value function $V(s,\mathcal{M})$ dependent on both state $s$ and MDP $\mathcal{M}$.
We further demonstrate that the true joint value function for a multi-scene environment, follows a multi-modal distribution which is not captured by traditional CNN / LSTM based critic networks. To this end, we propose a dynamic value estimation (DVE) technique, which approximates the true joint value function through a \emph{sparse} attention mechanism over multiple value function hypothesis / modes. The resulting agent not only shows significant improvements in the final reward score across a range of OpenAI ProcGen environments, but also exhibits enhanced navigation efficiency and provides an implicit mechanism for unsupervised state-space skill decomposition.

\end{abstract}

\section{Introduction}
\label{introduction}

While the field of reinforcement learning has shown tremendous progress in the recent years, generalization across variations in the environment dynamics remains out of reach for most state-of-the-art deep RL algorithms \cite{rajeswaran2017towards,zhang2018dissection,whiteson2011protecting}. In order to achieve the generalization objective, many deep RL approaches attempt to train agents on environments comprising of multiple levels or scenes from the same task \cite{wortsman2019learning,cobbe2019quantifying,zhu2017target,justesen2018illuminating,cobbe2019leveraging,kanagawa2019rogue,juliani2019obstacle}.  Although incorporating a wider source of data distribution in the training itself has shown promise in bridging the train and test performance, the inclusion of multiple scenes, each defined by a distinct underlying MDP, significantly increases the variance of samples collected for policy gradient computations \cite{cobbe2019leveraging,song2019empirical}.

The current approaches using multi-scene environments for training usually deal with the high variance problem by deploying multiple actors for collecting a larger and varied range of samples. For instance, \cite{zhu2017target,wortsman2019learning} use multiple asynchronous actor critic (A3C) models when training on the AI2-THOR framework  based visual navigation task. Similarly, \cite{cobbe2019leveraging,cobbe2019quantifying} deploy parallel workers to stabilize policy gradients for multi-level training on procedurally-generated game environments \cite{justesen2018illuminating}. While parallel sample collection helps in stabilizing the learning process, the obvious disadvantages of lower sample efficiency and higher hardware constraints, suggest the need for specialized variance reduction techniques in multi-scene RL.

Most RL generalization benchmarks \cite{nichol2018gotta,zhang2018study,cobbe2019quantifying,igl2019generalization} \emph{effectively} treat the collection of scenes as a \emph{single-MDP environment}. That is, a common and scene generic value function $V(s)$ is learned across all levels. By comparison, we propose an improved variance reduction formulation, which instead shows that the sample variance for a multi-scene environment is best minimized by acknowledging each scene as a separate MDP and then learning a joint value function $V(s,\mathcal{M})$ dependent on both state $s$ and MDP $\mathcal{M}$. 

However, given the lack of information about the operational level at train / test times, estimating the joint value function $V(s,\mathcal{M})$ presents a challenging problem.
To address this, we first show that the underlying true joint value function samples follow a multi-modal distribution. We then use this insight to propose a dynamic value estimation strategy, which approximates the overall value distribution through a progressively learned \emph{sparse} attention mechanism over the corresponding distribution modes. The sparse attention guided dynamic networks not only result in huge improvements in total reward on the OpenAI ProcGen benchmark, but also exhibit semantically desirable properties like enhanced navigation efficiency and provide an implicit framework for unsupervised state space decomposition.

The main contributions of this paper are summarized below. 
\begin{itemize}
\item \textbf{Enhanced Variance Reduction:} We propose an improved variance reduction formulation which shows that the sample variance for a multi-scene environment is best minimized by treating the each scene as a distinct MDP, and then learning a joint value function $V(s,\mathcal{M})$ dependent on both state $s$ and MDP $\mathcal{M}$.
\item \textbf{Clustering Hypothesis:} We show that the true scene-specific value function distribution is best described using a mixture model with multiple dominant modes, which are not fully captured by the current CNN or LSTM based critic networks.
\item \textbf{Novel Critic Module:} We propose a novel critic model which approximates the true multi-modal value function distribution through a \emph{sparse} attention mechanism over multiple value function hypothesis / modes.
\item \textbf{Implicit State-Space Decomposition:} We demonstrate that the learned sparse attention divides the overall state space into distinct sets of game skills. The collection of these skills represents a curriculum that the agent must master for effective game play.
\item \textbf{Navigation Efficiency:} Through both quantitative and qualitative results, we show that the sparse dynamic model leads to huge improvements in the navigation efficiency of the resulting agent. Furthermore, the high navigation efficiency of our method and its tendency to limit unnecessary exploration, presents an effective alternative to explicit reward-shaping \cite{zhu2017target, wortsman2019learning, laud2004theory,laud2003influence}, for penalizing longer episode-lengths / reward-horizons in multi-scene reinforcement learning.
\end{itemize}

\section{Problem Setup}

A typical multi-scene environment is characterized by a set of possible MDPs ${\mathcalbf{M}}: \{\mathcal{M}_1,\mathcal{M}_2, ... \mathcal{M}_N\}$, each defined by its own state space $\mathcal{S}_\mathcal{M}$, transition probabilities $P_\mathcal{M}(s_{t+1}|s_t,a_t)$, reward function $r_\mathcal{M}(s_{t},a_t,s_{t+1})$ and discount factor $\gamma$. An agent with action space $\mathcal{A}$ interacts with a randomly chosen and unknown MDP $\mathcal{M} \in \mathcalbf{M}$, to generate a trajectory $\tau: (s_0,a_0,s_1 ... s_T)$ with total reward $\mathcal{R}_\tau=\sum_{t=0}^{T-1} \gamma^{t} r_{\mathcal{M}}(s_{t},a_t,s_{t+1})$. The goal of the agent is to maximize the expected trajectory rewards over the entire MDP set $\mathcalbf{M}$, \emph{i.e.} 
$\mathbf{E}_{\tau,\mathcal{M}}\left[\mathcal{R}_{\tau,\mathcal{M}}\right]$.

\section{Motivation}
In this section, we present a step-by-step analysis which motivates the final method presented in Sec.~\ref{our_method}. We first begin by proposing an improved variance reduction formulation for multi-scene environments in Sec.~\ref{variance_reduction}. We then demonstrate the multi-modal nature of the joint value distribution in Sec.~\ref{clustering_hypothesis}. Finally we outline the main idea behind dynamic value estimation in Sec.~\ref{vf_error}.

\subsection{Variance Reduction for Multi-Scene Environments}
\label{variance_reduction}

For a single-MDP environment, with policy network $\pi$ (parameterized by $\theta$) and an action-value function $Q(s,a)$, the general expression for computing policy gradients with minimal possible sample variance can be written as, \cite{greensmith2004variance,schulman2015high},
\begin{equation}
\nabla_{\theta} J = \mathbf{E}_{s,a} \left[ (\nabla_{\theta} \log\pi(a|s)) \ \psi(s,a) \right], \label{eq:policy_grad_psi}
\end{equation}
where $\psi(s,a) = Q(s,a) - V(s)$ is the advantage function. Similarly for a multiple-scene environment, it can be shown that the optimal formulation for minimizing total sample variance is given by (please refer supp. material for proof),
\begin{equation}
\nabla_{\theta} J = \mathbf{E}_{s,a,\mathcal{M}} \left[ (\nabla_{\theta} \log\pi(a|s)) \ \psi(s,a,\mathcal{M}) \right], \label{eq:policy_grad_psi_multi}
\end{equation}
where $\psi(s,a,\mathcal{M}) = Q(s,a,\mathcal{M}) - V(s,\mathcal{M})$. Here $Q(s,a,\mathcal{M})$ and $V(s,\mathcal{M})$ represent the action-value and value function respectively for the particular MDP $\mathcal{M}$. However, since most of the times knowledge about the operational MDP $\mathcal{M}$ is unknown to the agent, the current policy gradient methods continue to use a single scene-generic value function estimate $\hat{V}(s)$ for variance reduction. However, $\hat{V}(s)$ then is essentially an estimate of the global average over the underlying scene-specific value functions $\{V_{\mathcal{M}_1}(s),V_{\mathcal{M}_2}(s), ... , V_{\mathcal{M}_N}(s)\}$, and thus gives a poor approximation of the joint value function for a given MDP.

We next show that such a simplification is not necessary and present an approach for obtaining a better approximation for the joint value function $V(s,\mathcal{M})$.

\begin{figure*}[ht]
\floatbox[{\capbeside\thisfloatsetup{capbesideposition={right,center},capbesidewidth=5cm}}]{figure}[\FBwidth]
{\caption{\textbf{Clustering Hypothesis.} Column 1-2 demonstrate multi-modal nature of the true value function distribution, for an intermediate policy $\pi$, on a set of randomly selected 50 levels from the CoinRun environment. The true value estimate for a state image shown in Column 1 can be characterized by one of the many clusters. Column 3: In contrast, the LSTM based value predictions, though showing some variance with MDP $\mathcal{M}$, fail to capture the multiple dominant modes exhibited by the true value function distribution.}\label{fig:clustering_hypothesis}}
{\includegraphics[width=11cm]{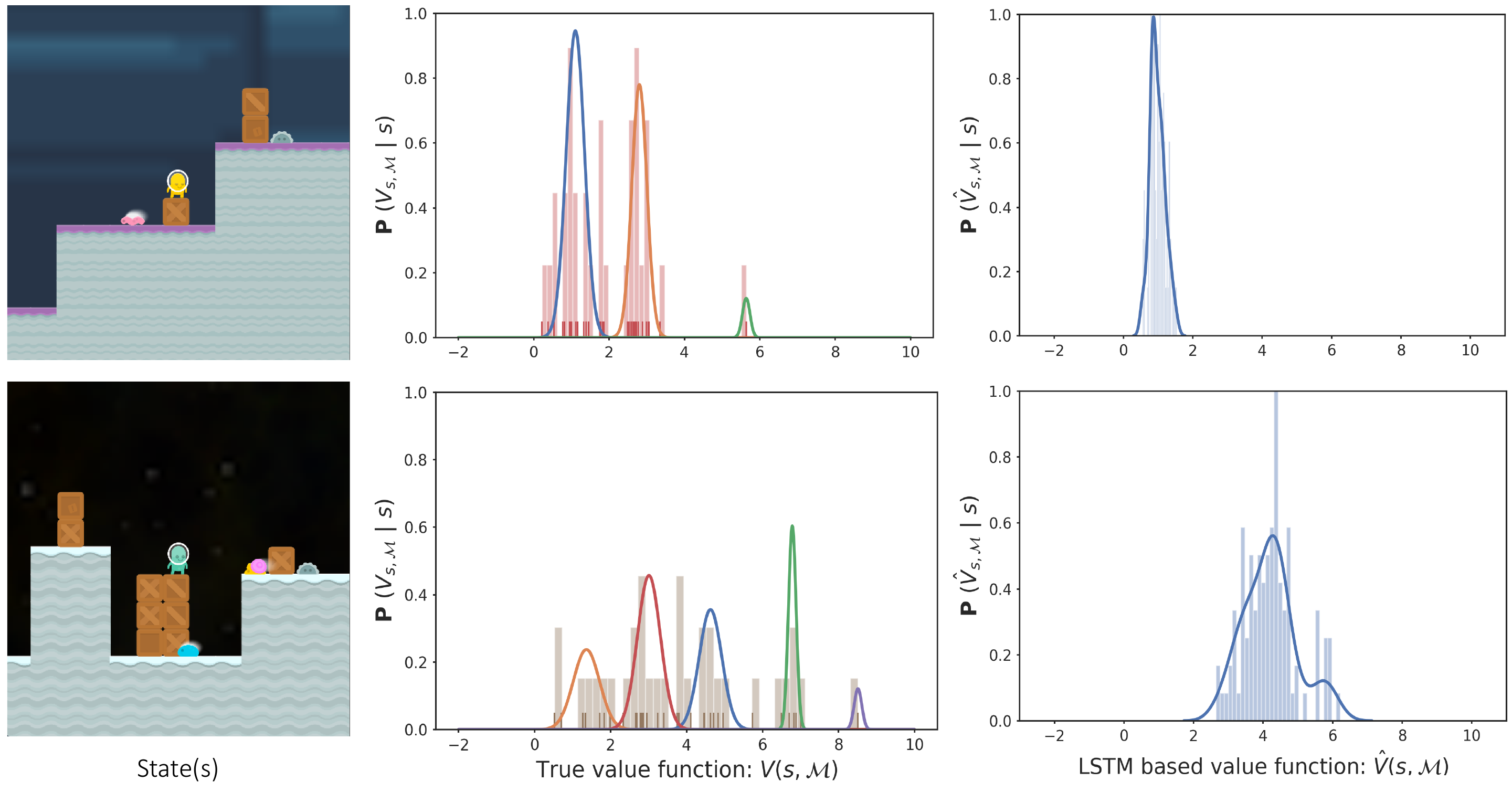}}
\end{figure*}

\subsection{Clustering Hypothesis}
\label{clustering_hypothesis}
Training on multi-scene environments over the same domain task can lead to ambiguity in value function estimation. That is, two visually similar states could have very different value function estimates corresponding to distinct scenes / levels.
In this section, we empirically demonstrate that unlike a single-scene environment, the true value function for a multi-scene environment (having scenes with similar state spaces), is better described by a multi-modal distribution.

 \textbf{Empirical Demonstration on OpenAI CoinRun\footnote{Please refer supplementary materials for further details.}.} To test the above hypothesis, we finetune separate critic networks over a fixed policy $\pi$, to obtain the true MDP-specific value function estimates $\{V(s,\mathcal{M}_1),V(s,\mathcal{M}_2), ... ,V(s,\mathcal{M}_{50})\}$ corresponding to a random selection of 50 levels from the CoinRun ProcGen environment \cite{cobbe2019quantifying}. We then use a Gaussian Mixture Model (GMM) for fitting these $V(s,\mathcal{M}_i) \{i \in [1,50]\}$ samples. Results are shown in Fig. \ref{fig:clustering_hypothesis}. We clearly observe that the true value function estimates form a multi-modal distribution that is not captured by traditional CNN or LSTM based critic networks.

\subsection{Minimizing Value Prediction Error}
\label{vf_error}

\textbf{Theorem 1.} The sample variance ($\nu$) for policy gradients defined by Eq. \ref{eq:policy_grad_psi_multi}, can be minimized by reducing the prediction error $\epsilon$ between the true joint value function $V(s,\mathcal{M})$ and the predicted estimate $\hat{V}(s,\mathcal{M})$, where
\begin{equation}
    \epsilon = \mathbf{E}_{s,\mathcal{M}} \left[\left(V(s,\mathcal{M}) - \hat{V}(s,\mathcal{M})\right)^2\right].
\end{equation}

The proof is provided in supp. materials. We now use the insight provided by Theorem 1 to propose the following solution for reducing the policy gradient sample variance.

\begin{figure}[ht]
\begin{center}
\centerline{\includegraphics[width=0.9\columnwidth]{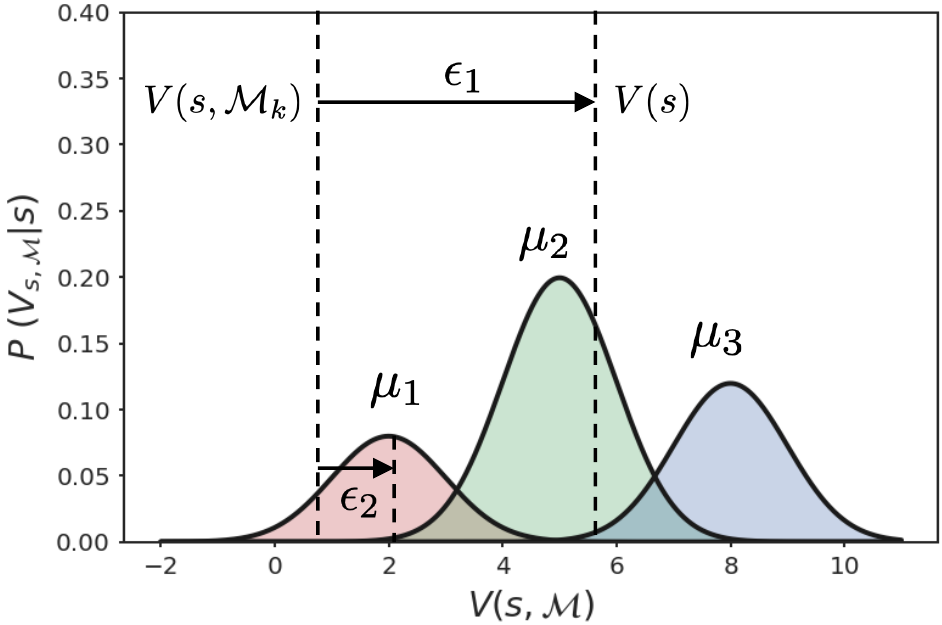}}
 \caption{\textbf{Proposed Approach.} Traditional methods learn a single scene-generic value function estimate $V(s)$ which leads to high prediction error ($\epsilon_1 > \epsilon_2$). In contrast we propose that the overall prediction error can be significantly reduced by approximating the joint value function $V(s,\mathcal{M})$ as the mean of the nearest cluster.}
\label{fig:gen_idea}
\end{center}
\vskip -0.4in
\end{figure}

\textbf{Proposed Solution:} While the exact estimation of the true value function is infeasible without knowledge of MDP $\mathcal{M}$, we use the results of Section \ref{clustering_hypothesis}, to assert that the prediction error can be reduced by approximating the value function as the mean value of the cluster to which the current MDP belongs. Fig.~\ref{fig:gen_idea} provides an overview of this idea.

\section{Our method}
\label{our_method}
The solution proposed in Sec.~\ref{vf_error} can be implemented through a sparse attention mechanism over the value function modes $\mu_i$, wherein the attention parameters are $1$ for the closest cluster and $0$ otherwise. However the sparse attention parameters are not fixed, as the cluster to which an MDP belongs is expected to change while training the RL agent. To address this, we first model the predicted value function through a generalized attention mechanism over the value function modes (Sec.~\ref{dynamic_vf_estimation}) and then propose a novel loss function which progressively enforces sparse attention parameters based on the training dynamics (Sec.~\ref{sparse_loss}).

\subsection{Continuous Dynamic Estimation Formalism}
\label{dynamic_vf_estimation}
Mathematically, given that the \emph{true} value function follows a Gaussian Mixture Model (GMM) like,
\begin{equation}
P \left(V(s_t,\mathcal{M}) \vert s_t \right) = \sum_{i=1}^{N_b} p_i \ \mathcal{N}(V(s_t,\mathcal{M})|\mu_i,\sigma_i^2),
\label{eq:gmm_model}
\end{equation}
we propose to model the \emph{predicted} value function as,
\begin{multline}
\hat{V}(s_t,\mathcal{M}) = \sum_{i=1}^{N_b} \alpha_i(s_t, \mathcal{M}) \ \mu_i(s_t),\\
s.t. \quad \alpha_i > 0, \quad \sum_i^{N_b} \alpha_i=1.
\label{eq:gmm_mean}
\end{multline}
That is, given a state $s_t$, we predict $N_b$ distinct value function hypotheses $\{\mu_1(s),\mu_2(s), ... , \mu_{N_b}(s)\}$ (one corresponding to each cluster). The final value prediction is then modelled as the weighted combination of these value hypotheses using attention parameters $\alpha_i$. Wherein, the attention parameters $\alpha_i(s_t, \mathcal{M})$ are used to capture the similarity between the $i^{th}$ value hypothesis and the true value for MDP $\mathcal{M}$. In practice, since the current MDP $\mathcal{M}$ is not known, the parameters $\alpha_i$ are learned from (state, episode trajectory) pairs $\{s_t,\tau^{t-}\}$ ($\tau^{t-} : \{s_0,a_0, .... s_{t-1}\}$ is the trajectory till time $t-1$), using a Long Short Term Memory (LSTM) network.

\subsection{Enforcing Sparse Attention}
\label{sparse_loss}

We first define two metrics to describe the attention parameter distribution, \emph{confusion} and \emph{contribution}.  Confusion ($\delta$) is a measure of uncertainty as to which cluster, the current state-trajectory pair $\{s_t,\tau^{t-}\}$ belongs to. On the other hand, contribution ($\rho$), as the name suggests, determines the `contribution' of a cluster in the overall value function estimation across  a general trajectory sequence $\tau: \{s_0,a_0,s_1,a_1, ... s_T\}$. Formally, confusion and contribution are defined as,
\begin{align}
    \delta(s_t,\tau^{t-}) = \frac{1}{N_b . \sum_i \alpha^2_i(s_t,\tau^{t-})}, \label{eq:conf}\\
    \rho_i(\tau) = \frac{1}{T} \sum_{t=1}^{T} \delta(s_t,\tau^{t-}) \  \alpha_i(s_t,\tau^{t-}).
    \label{eq:contrib}
\end{align}

We now note that an increase in sparsity of cluster assignments $\{\alpha_1,\alpha_2, \dots \alpha_{N_b}\}$ is equivalent to maximization of their $l_2$ norm. Thus using Eq. \ref{eq:conf}, it corresponds to minimization of the confusion ($\delta$) metric. However, a mere enforcement of sparsity may encourage convergence to solutions where only one of the clusters is active. We also want to ensure that each cluster contributes equally in the $(s,\mathcal{M})$ space. To achieve this, we propose the following \emph{confusion-contribution loss},

\begin{multline}
L^{CC} = k_1 \ \mathbf{E}_{s_t,\tau^{t-}} \left[\log \delta(s_t,\tau^{t-})\right] + \\ k_2 \ \mathbf{E}_{\tau}  \left[\log \left(\sum_i^{N_b} \rho^2_i(\tau)\right)\right]. \label{eq:l_cc}
\end{multline}

We emphasize that the state space must have already been well explored\footnote{For OpenAI ProcGen, we consider state space to be sufficiently explored when the avg. episode length plateaus / starts decreasing.} by the agent, prior to the application of confusion-contribution loss. If applied prematurely, due to the continuous nature of neural networks, the sparse cluster assignment is incorrectly generalized across the entire state space. This would lead to a detrimental impact on value function estimation for the currently unexplored states. Also, such a mistake is hard to recover from, because for any state $s \in \mathcal{S}$, the sparse assignment ensures that the gradients for all but one cluster are approximately zero.

\section{Related Work}
\label{related_work}
\textbf{Meta Reinforcement Learning.}
\cite{duan2016rl,wang2016learning} previously proposed the use of recurrent neural networks and episode trajectories as a meta-RL approach for adapting to environment dynamics. While in theory, an LSTM is capable of learning multi-modal distributions, we find that in practice the vanilla-LSTM based conditional value function distribution (for a given state) is usually characterized by a single dominant mode (refer Fig. \ref{fig:clustering_hypothesis}), and thus fails to capture the multi-modal nature of true value function distribution. In contrast, our method explicitly forces multiple dominant modes while estimating the cluster means $\mu_i$ and uses episode trajectories to compute the assignment $(\alpha_i)$ of the current state sample to each cluster\footnote{In this paper, we occasionally use the term cluster to refer to the value function hypothesis, since each hypothesis represents the mean of a distinct cluster in the true value distribution.}.

\textbf{Recurrent Independent Mechanisms.} \cite{goyal2019recurrent} propose modular structures called Recurrent Independent Mechanisms (RIMs) which specialize to different dynamic processes within an environment, and communicate sparingly through a sparse attention mechanism. While RIMs and our method share the idea of having sparse attention, our work differs significantly as RIMs use separate recurrent models with independent dynamics, whereas we deploy a single actor-critic network with shared dynamics.
Another important distinction is that \cite{goyal2019recurrent} assume a sparse structure from the beginning and hence require environments with clearly independent dynamic processes. In contrast, the sparse assignment in our method is learned progressively and only after sufficient exploration, which allows for a more informed division of the state space (Sec.~\ref{cluster_state_analysis}).

\textbf{Generalization in Reinforcement Learning.} Recently, multi-scene environments have been extensively used to study and address the problem of overfitting in RL. 
\cite{cobbe2019quantifying} deploy standard regularization techniques from supervised learning like dropout, batch-normalization, L2-regularization to counter overfitting when training on the multi-scene CoinRun environment. \cite{rajeswaran2016epopt} learn robust policies over an ensemble of scenes by formulating the overall objective as the expected reward over scenes with the worst performance. Noise injection techniques like \cite{igl2019generalization,tobin2017domain} add noise to the model parameters in order to improve the generalization capability.  While effective, these works are seen to reduce overfitting at some expense to the training performance. Our work is different as it doesn't focus on generalization specifically, but improves both training and test time performance by learning a better value function estimate. Furthermore, we note that our approach is orthogonal to these works and would likely also benefit from use of regularization methods.

\textbf{Distributional RL.} Recent works like \cite{bellemare2017distributional,dabney2018distributional,choi2019distributional} aim to directly learn the value function distribution instead of modelling the expected return. Our work differs in the following aspects. First, distributional RL methods need to discretize the return space using a high number of support locations/nodes (\emph{e.g.}~$N=200$ for \cite{bellemare2017distributional}) to approximate the overall value distribution. In contrast, we approximate the joint value function through only the modes of the underlying distribution and thus require very few output nodes ($N \in [2,5]$). Second, the current work on distributional RL is limited to off-policy methods with a shared replay buffer. The use of a large replay buffer implies that the overall sample distribution changes minimally from one update to another. This is in sharp contrast to the high variance seen in on-policy training which is the focus of current work. To the best of our knowledge, our work is the first stable method for approximating value distribution in on-policy RL.

\textbf{Bootstrapped DQN.} \cite{osband2016deep} propose the use of multi-head Q-networks for bootstrapped DQNs. However, they aim to facilitate deep exploration and hence select the Q-network head for a given episode randomly. In contrast, we aim to reduce the prediction error with the true value function distribution and also present a novel approach which progressively learns the most representative cluster for each state $s \in \mathcal{S}$ through the confusion-contribution loss.

In addition to the above, we note that \cite{liang2018vmav} propose an attention-based value function network for model-based reinforcement learning. However, they consider an attention over past trajectory states, whereas we consider an attention over multiple value function hypotheses.
\section{Evaluation on OpenAI Procgen}
\label{procgen}

\begin{figure*}[ht]
\floatbox[{\capbeside\thisfloatsetup{capbesideposition={right,center},capbesidewidth=5cm}}]{figure}[\FBwidth]
{\caption{\textbf{Learning curves} for LSTM-PPO, sparse and non-sparse dynamic models, illustrating differences in sample complexity, total reward and episode lengths. We clearly observe that our method (with sparse attention) achieves higher total rewards (top-row) while needing much shorter episode lengths (bottom-row) per reward unit.}\label{fig:procgen_train_curves}}
{\includegraphics[width=11cm]{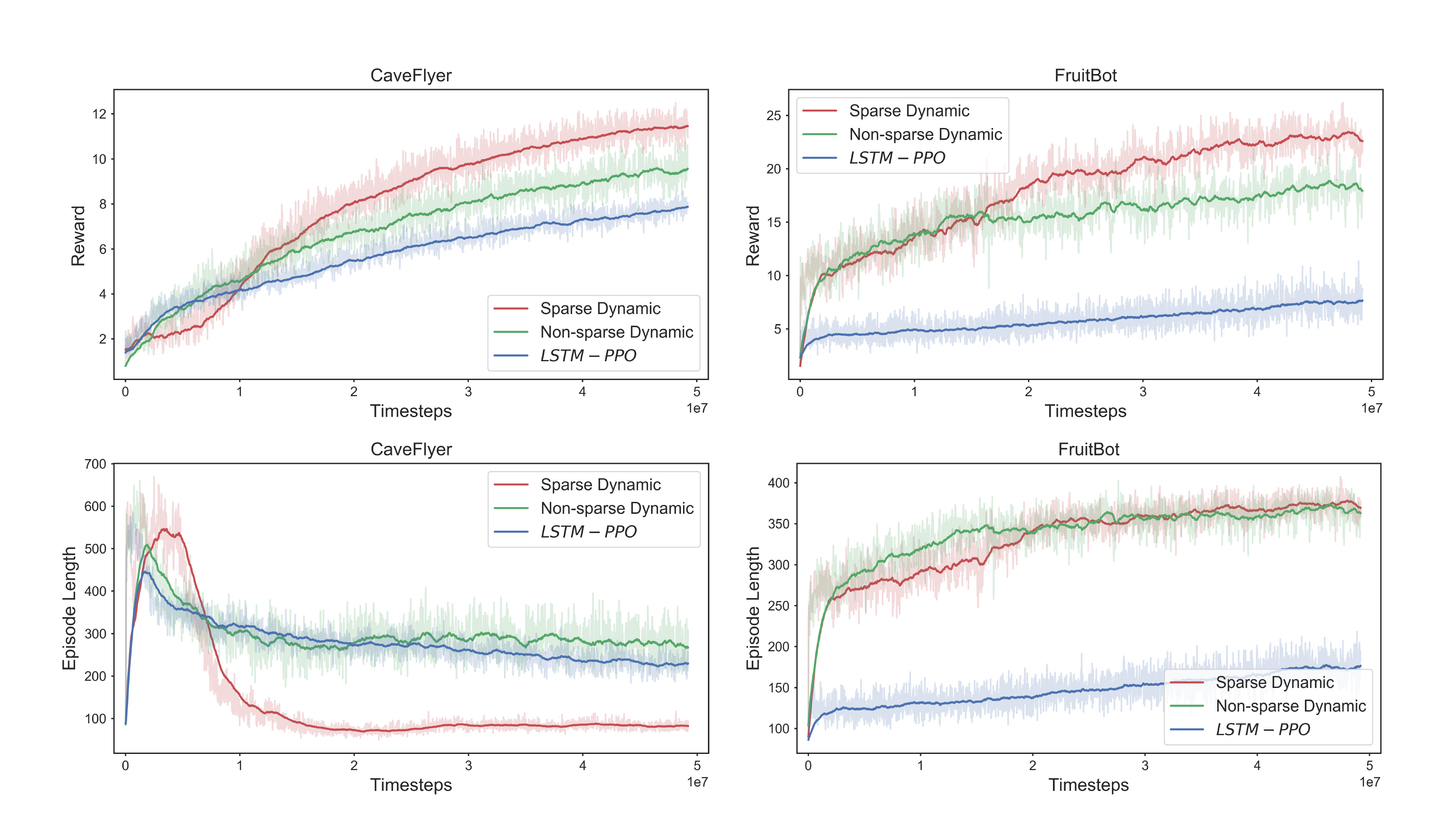}}
\end{figure*}

\begin{table*}[!htbp]
  \vskip 0.1in
  \centering
  \begin{tabular}{c|l|ccc|ccc}
    \toprule
    \multicolumn{2}{c}{} & \multicolumn{3}{c}{Total Reward}  & \multicolumn{3}{p{6cm}}{\centering Navigation Efficiency [$\times 10^{-2}$] \\ (Total reward / Episode length)}\\
    \midrule
    Class & Environment & PPO & {\centering DVE (Ours)} & Sparse DVE (Ours) & PPO & DVE (Ours) & Sparse DVE (Ours) \\
    \midrule
    \multirow{4}{*}{Class 1} & CoinRun  & 7.75 & 9.16 & \textbf{9.62} & 6.15 & 11.59 & \textbf{15.48} \\
    & CaveFlyer  & 6.82 & 9.02 & \textbf{11.57}  & 3.02 & 3.27 & \textbf{15.39}\\
    & Climber & 7.50 & 8.14 & \textbf{10.17} & 4.21	& 3.59	& \textbf{5.92} \\
    & Jumper  & 6.61 & 6.52 & \textbf{6.65} & 2.80 & 3.61 & \textbf{8.43}\\
    \midrule
    \multirow{5}{*}{Class 2} & Plunder  & 7.13 & 17.16 & \textbf{18.42} & 1.44	& 2.19 & \textbf{2.48}\\
    & DodgeBall & 10.98 & 11.25 & \textbf{12.76} & 2.73 & 2.44 & \textbf{4.45} \\
    & FruitBot  & 7.33 & 18.32 & \textbf{23.08} & 4.25 & 5.02 & \textbf{6.17}\\
    & StarPilot & 17.94 & 18.08 & \textbf{19.81} & 5.48 & 5.28 & \textbf{6.19} \\
    \bottomrule
  \end{tabular}
  \caption{\textbf{Final performance comparison}. We clearly see that the sparse dynamic model results in huge gains in both the total reward as well as the overall navigation efficiency \emph{i.e.}~it achieves higher rewards while needing much shorter episode lengths (per reward unit).}
  \label{tb:procgen_results}
\end{table*}

\subsection{Training Configurations}
\textbf{CNN-LSTM Baseline.} The baseline model closely follows\textsuperscript{\ref{fn:lstm_baseline}} the one described in \cite{cobbe2019leveraging}. Both actor and critic share an IMPALA-CNN network \cite{espeholt2018impala} modeled in the form of an LSTM \cite{hochreiter1997long}. The output from the LSTM is then fed to both actor and critic separately, and is followed by a single fully-connected (FC) layer to compute the action scores and value function, respectively. The choice of hyper parameters is kept similar to that described in \cite{cobbe2019leveraging} in order to allow for comparison on the baseline values \footnote{\label{fn:lstm_baseline} \cite{cobbe2019leveraging} use an IMPALA CNN network as the baseline, while we use an IMPALA CNN-LSTM network.}, though, local hyper-parameter search was performed for each game to achieve the best average reward score.

\textbf{Dynamic.} The dynamic model is quite similar to the baseline and only requires few changes in the critic network to model the dynamic estimate described by Eq.~\ref{eq:gmm_mean}. The attention parameters $\alpha_i(s_t,\tau^{t-})$ are modeled using a fully connected layer followed by a softmax function, while the mean estimates $\mu_i(s_t)$ are learned using a single FC layer. The number of dynamic clusters/nodes $N_b$ is treated as a hyperparameter\footnote{The Bayesian optimal choice for the number of clusters $N_b$ can be determined by minimizing the Akaike Information Criterion (AIC) \cite{akaike1987factor,forster2011aic} for the learned value function distribution. We demonstrate this in supp. material.}.
For most ProcGen environments, the optimal choice for $N_b$ is found to lie within the range $[2,5]$.

\textbf{Sparse Dynamic.} The only difference with the original dynamic model is the additional application of the confusion-contribution loss $(L_{CC})$ during the training process. A hyperparameter selection of $\{k_1=0.1, k_2=1\}$ was seen to give competitive results for most ProcGen environments.

All the three configurations are trained using the Proximal policy optimization (PPO) \cite{schulman2017proximal} algorithm, which is ran with 4 parallel workers for gradient computations as this is seen to enhance performance. Each worker is trained for 50M steps, thus equating to a total of 200M steps across all the 4 workers. All results are reported as the average across 4 runs using 500 levels for training.

\subsection{Test Environments}
We test our method on 8 OpenAI ProcGen \cite{cobbe2019leveraging} environments: Dodgeball, FruitBot, StarPilot, CoinRun, CaveFlyer, Climber, Jumper and Plunder. Note that each game is characterized by a different rate of state exploration. Thus, depending upon the type of environment, we adopt the following strategies for applying the sparse loss.

\textbf{Class-1.} For games allowing rapid state space exploration at the beginning, the confusion-contribution loss can be applied quite early to promote sparsity. In fact, because the policy gradient and value function loss dominate the initial training updates, we apply the confusion-contribution loss from the start. CoinRun, CaveFlyer, Climber and Jumper belong to this set and are labelled as \emph{class-1} environments.

\textbf{Class-2.} In contrast, other games display a much more gradual expansion of explored state space, exhibiting a positive correlation between episode lengths and the total reward. The sparse loss for such environments can only be applied after the rate of increase of average episode length has declined. Thus, the application of confusion-contribution loss is usually preceded by pre-training\footnote{For maintaining consistency, a pre-training procedure same as the one described above, is followed for all training configurations on class-2 environments.}
 with the original dynamic model for 50M timesteps (per worker). Games like Plunder, Dodgeball, FruitBot and StarPilot are part of this set and are labelled as \emph{class-2} environments.

\begin{figure*}[ht]
\vskip 0.1in
\begin{center}
\centerline{\includegraphics[width=0.8\linewidth]{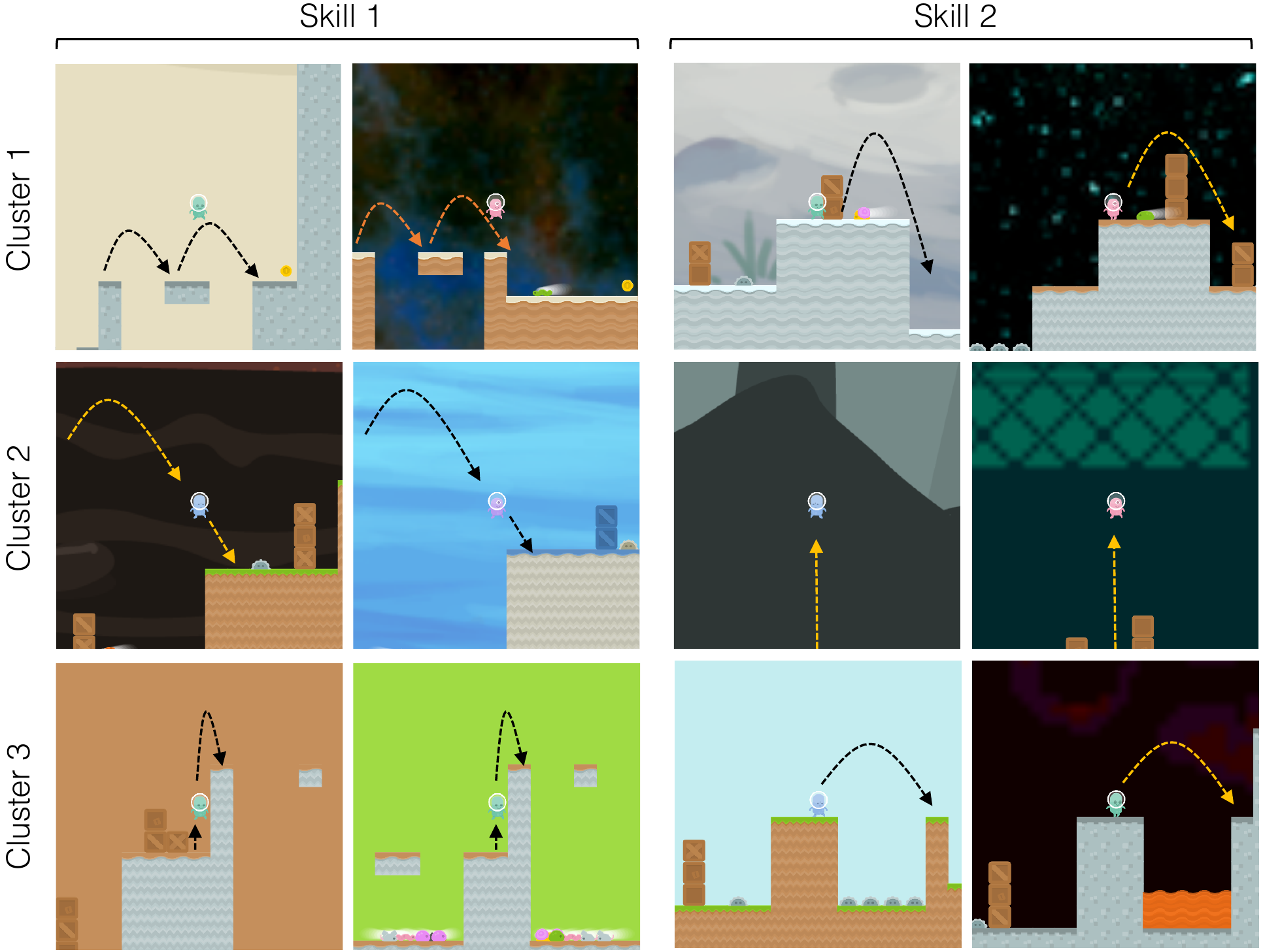}}
\caption{\textbf{Visualizing Cluster Features.} Examples of key obstacles types learned by each cluster in the CoinRun Environment. We note that the sparse training divides the overall state space into a distinct sets of game skills.}
\label{fig:coinrun_cluster_feats}
\end{center}
\vskip -0.2in
\end{figure*}

\section{Results}

\textbf{Total Reward and Sample Complexity.}
Table~\ref{tb:procgen_results} reports our results on various OpenAI-ProcGen environments.
We clearly see that the use of dynamic value estimation (with sparse attention) leads to huge improvements in both the final reward and sample complexity (refer Fig.~\ref{fig:procgen_train_curves}) over the baseline model. For instance,  our method (with sparse attention) results in an increase of 69.7\%, 158.3\% \& 215.1\%  in the average episode reward, for the CaveFlyer, Plunder and FruitBot environments respectively. 

\textbf{Navigation Efficiency.} In addition to reporting results for average episode reward, we also compare the model performance based on the agent's efficiency in completing a game level. The navigation efficiency is thus measured by the ratio of the final reward and average episode length. Results are reported in Table~\ref{tb:procgen_results}. We clearly see that the sparse model leads to better reward scores while on average, using much fewer timesteps per episode\footnote{Note that the ProcGen environments have no explicit penalty for discouraging longer episode lengths.}. For instance, we see that the sparse dynamic model leads to an increase of 151.7\%, 409.1\% \& 200\% in the reported navigation efficiency scores, for the CoinRun, CaveFlyer, and Jumper environments respectively. We discuss more on this phenomenon in Sec.~\ref{navigation_efficiency}.

\textbf{Requirement for sufficient state space exploration.} Interestingly, we also see that a saturation in the rate of state space exploration is necessary for getting gains with the sparse model. This is illustrated through the training curves for the game of FruitBot (refer Fig.~\ref{fig:procgen_train_curves}), where relative gains over the non-sparse dynamic model occur only after a decline in the rate of increase of average episode length. 

\section{Analysis}
\subsection{Learning Implicit State Space Decomposition}
\label{cluster_state_analysis}

A key advantage of our method can be seen in its ability to achieve an unsupervised division of the state space into distinct sets of game skills. The state space decomposition is achieved through the sparse cluster assignments, wherein the network learns to assign each state, trajectory pair $\{s_t,\tau^{t-}\}$ to a distinct value cluster. In this section, we use this sparse property of our method to visualize different obstacle types characteristic of each cluster in the CoinRun Environment.

To visualize the distinguishing features for each cluster, we first extract the set of states $\mathcal{S}_i$ for which each cluster is active. The latent representations (output of the LSTM network) for these states are used to map each $s \in \mathcal{S}_i$ to a two dimensional embedding space using TSNE \cite{maaten2008visualizing}. This embedding is then manually analysed for clusters to then identify the salient obstacle classes. 

Fig.~\ref{fig:coinrun_cluster_feats} shows some key obstacle types for each cluster. We observe that each cluster is responsible for predicting the value function on a distinct set of obstacles/skills. For instance, \emph{Cluster-1} is responsible for value estimation in cases like double-jump from one side to another (Skill-1) and crossing over moving enemies (Skill-2). On the other hand, \emph{Cluster-2} handles landing after jumps from higher ground (Skill-1) and high jumps with very limited visibility of coming obstacles (Skill-2). Finally, \emph{Cluster-3} takes care of precision climbs (Skill-1) and jumps over wide valleys (Skill-2).

Thus, we see that each disjoint state space set $\mathcal{S}_i, \ i\in[1,N_b]$ represents a distinct curriculum of game skills that must be learned for mastering the overall multi-scene game environment. This division is semantically desirable and is analogous to the human learning paradigm wherein it is quite common to break down a complex task into a set of manageable skills before attempting the complete task. 

\begin{figure*}[ht]
\vskip 0.2in
\begin{center}
\centerline{\includegraphics[width=0.9\linewidth]{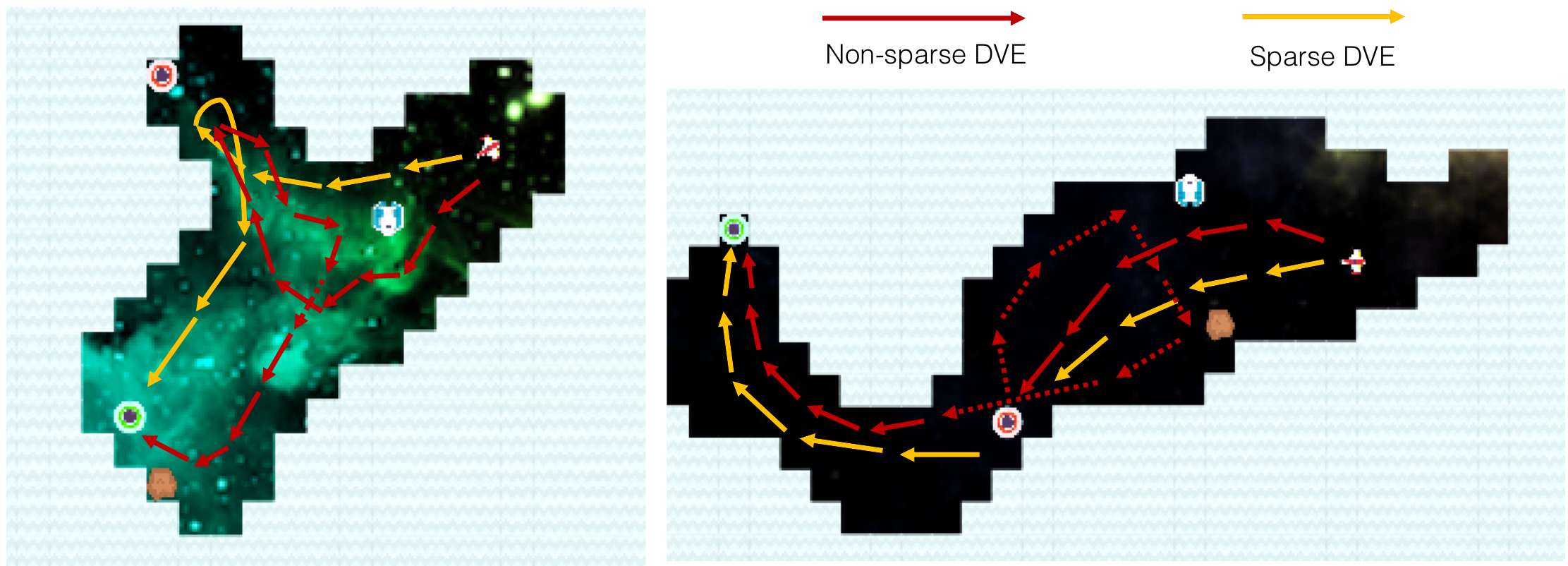}}
\caption{\textbf{Comparing Navigation Efficiency.} Demonstrating qualitative difference between trajectories for sparse and non-sparse dynamic agents. Our method shows higher efficiency in navigating to the final goals (red \& green spheres).}
\label{fig:trajectories}
\end{center}
\vskip -0.3in
\end{figure*}

\subsection{Navigation Efficiency}
\label{navigation_efficiency}

A peculiar feature arising as a result of learning sparse attention parameters (over non-sparse dynamic model) can be seen in terms of the highly improved \emph{navigation efficiency}. That is, the agent on average uses far fewer time-steps per episode while achieving similar or better reward score. This massive difference in time-steps results from two reasons.
\begin{itemize}
    \item The tendency to use fewer time-steps is a direct consequence of optimizing the discounted reward function with $\gamma < 1$ \cite{schulman2015high}. As a result, the agent is incentivized to minimize the number of steps between the current state and the next reward. Hence, more accurate policy updates (with lower sample variance) should lead to fewer timesteps.
    \item As explained in Sec.~\ref{sparse_loss}, the expansion in state space after application of the confusion-contribution loss can lead to potential errors in value function estimation. Thus, the sparse dynamic agent learns to maximize the utilization of the already explored state space. 
\end{itemize}

In this section, we will qualitatively analyse how learning a more accurate value function leads to shorter (and efficient) episode trajectories for the sparse agent. Note that the computation of a suboptimal value function at a \emph{critical} environment state (\emph{e.g.}~a tricky obstacle) can cause the agent to underestimate the \emph{advantage} of choosing an action which leads to a faster route to the final destination / goal. We next try to identify these \emph{critical} states by comparing episode trajectories for the sparse and non-sparse dynamic agents on the CaveFlyer environment.

\textbf{Game Description.} The goal of the Caveflyer environment (refer Fig. \ref{fig:trajectories}) is to destroy the red spheres and finally reach the green sphere while avoiding intermediate obstacles. The agent receives a small reward of +3 on destroying a red sphere and an end of episode reward of +10 on successfully reaching the green one. Direct collisions with an obstacle or the red sphere cause immediate episode termination.

Trajectories for both sparse and non-sparse dynamic agents are shown in Fig.~\ref{fig:trajectories}. We see that the non-sparse agent after destruction of the red sphere (critical state), effectively restarts its search for the next target, while often revisiting already encountered states. In contrast, the sparse agent with its more accurate value estimates, realizes that the expected value for exploring unseen parts of the cave is much higher than revisiting previous states. Doing so not only helps the sparse agent reach the end goals much faster, but also eliminates the need for evading obstacles that it has already crossed (thereby increasing the episode success rate).

We also note that, the balance between the sparse model's reluctance towards state space expansion and maximization of total reward can be modulated through the scale of coefficients of the confusion-contribution loss. In this regard, the high navigation efficiency of our method provides an effective alternative to designing explicit reward shaping penalties \cite{zhu2017target,wortsman2019learning,laud2004theory} for promoting reduced episode lengths.

\section{Conclusion}
This paper introduces a novel dynamic value estimation strategy for enhanced variance reduction in multi-scene reinforcement learning environments. The proposed approach significantly outperforms the baseline model on a range of OpenAI ProcGen environments, while exhibiting much higher navigation efficiency to complete a game level. Additionally, we observe that the learned sparse attention parameters divide the overall state space into disjoint subsets. We show that each subset focuses on a distinct set of game-skills, which is semantically desirable and draws a strong parallel with the human learning paradigm.

\newpage
\bibliography{references}
\bibliographystyle{icml2021}

\newpage

\onecolumn
\newgeometry{
    textheight=9.0in,
    textwidth=5.7in,
    top=1in,
    headheight=12pt,
    headsep=25pt,
    footskip=30pt
  }
\fancyfootoffset{0pt}
\setlength\topmargin{-24.95781pt}

\begin{appendices}
\section{Additional Evaluation on the Visual Navigation Task}
\label{visnav}

The proposed dynamic value estimation strategy (with sparse attention) is highly effective and widely generalizable to any task domain utilizing multi-scene environments. In this section, we demonstrate its further efficacy on the domain of visual navigation, which heavily relies on using multiple-scene environments for achieving better generalization performance. 

\subsection{Task Definition}
The task of visual navigation consists of a set of scenes $\mathcalbf{S} = \{\mathcal{S}_1, \mathcal{S}_2, ... \mathcal{S}_n\}$, and a set of possible object classes $\mathcalbf{O} = \{\mathcal{O}_1,\mathcal{O}_2 .. \mathcal{O}_m\}$. Note that the set $\mathcalbf{S}$ is just another annotation for the MDP set $\mathcalbf{M} = \{\mathcal{M}_1, \mathcal{M}_2, ... \mathcal{M}_n\}$, as each scene  $\mathcal{S}_i$ represents a distinct underlying MDP and is characterized by a different room setup, distribution of objects and lighting conditions.

\textbf{Navigation Task Description.} A single navigation task $\mathcal{T}$ consists of an agent with action space $\mathcal{A}$ situated at a random starting position $p$ in the one of the scenes $\mathcal{S}_i$. The goal of the agent is to reach an instance of the target class $\mathcal{O}_k$ (given as a Glove embedding \cite{pennington2014glove}) within a certain number of steps. The agent then continues interacting with the environment using a policy $\pi_\theta$ until it chooses a termination action. The episode is considered a success, if and only if, at the time of termination, the target object is sufficiently close and in the field of view of the agent.

\subsection{Experimental design}



The performance on the visual navigation task is measured using the following evaluation metrics,
\begin{itemize}
    \item \textbf{SPL.} Success weighted by Path Length (SPL) \cite{anderson2018evaluation} provides a measure for the navigation efficiency of the agent and is given by $\frac{1}{N} \sum_{i=1}^N S_i \frac{L_i}{\max(P_i,L_i)}$.
    \item \textbf{Success rate.} The average rate of success: $\frac{1}{N} \sum_{i=1}^N S_i$.
    \item \textbf{Total reward.} Average total episode reward: $\frac{1}{N} \sum_{i=1}^N \mathcal{R}_i$.
\end{itemize}




In above, $N$ is the number of episodes, $S_i \in \{0,1\}$ indicates the success of an episode, $P_i$ is the path length, $L_i$ is the optimal path length to any instance of the target object class in that scene, and $\mathcal{R}_i$ is the episode reward.
The baseline results are reported using the non-adaptive and self-adapting (SAVN) A3C models from \cite{wortsman2019learning} with 12 asynchronous workers. We then modify the critic network as per Sec.~4 of the main paper to get the dynamic training configuration for both baselines. 


The agent is trained using the  AI2-THOR environment \cite{kolve2017ai2} which consists of 120 distinct scenes. A train/val/test split of 80:20:20 is used for selecting the best model based on highest success rate. Note that the final results are reported after 5M episodes of training which equates to $\approx$ 96 hours of training time on 2 Nvidia RTX 2080 Ti GPUs.

\subsection{Results}


As described in Table \ref{tab:visnav}, the dynamic A3C model results in significant improvements across all 3 performance metrics for both self-adaptive and non-adaptive baselines. Furthermore, we note that the dynamic model provides a huge boost in sample efficiency of the navigation agent (refer Fig.~\ref{fig:visnav_train_curve}). For instance, the dynamic nonadaptive A3C model reaches a test success rate of $\approx 30$\% after only 2M episodes, which in contrast with the non-adaptive baseline takes around 4M episodes. Similarly, the dynamic model for self-adaptive A3C achieves a test success rate of $\approx 40$\% after only 1M episodes while the corresponding baseline has a success rate of only $30.8$\%. Given the huge amounts of training time required for this task, the dynamic model promises to be a real asset for applications with much stricter time constraints. For instance, the high sample efficiency of our method would be highly useful for quick fine-tuning in mobile navigation robots operating in the real world \cite{chancan2019visual}.

\begin{table}
    \RawFloats
	\begin{minipage}{0.55\linewidth}
        \begin{center}
        \begin{small}
        \begin{tabular}{lcccl}
        \toprule \midrule
        Method & SPL & Success & Total Reward \\
        \midrule \midrule 
        A3C  & 14.3 & 31.8 & 1.413 \\
        Dynamic A3C & \textbf{15.4} &  \textbf{36.5} & \textbf{1.638}\\
        \midrule 
        SAVN & \textbf{15.19} & 37.1 & 1.652 \\
        Dynamic SAVN & 14.81 &  \textbf{38.7} & \textbf{1.824}\\
        \bottomrule
        \end{tabular}
        \end{small}
        \end{center}
        \vskip 0.15in
        \caption{Comparison on key evaluation metrics for visual navigation. We observe improvements when using the dynamic critic network on both the non-adaptive and adaptive baselines.}
        \label{tab:visnav}
	\end{minipage}\hfill
	\begin{minipage}{0.35\linewidth}
		\centering
		\includegraphics[width=\linewidth]{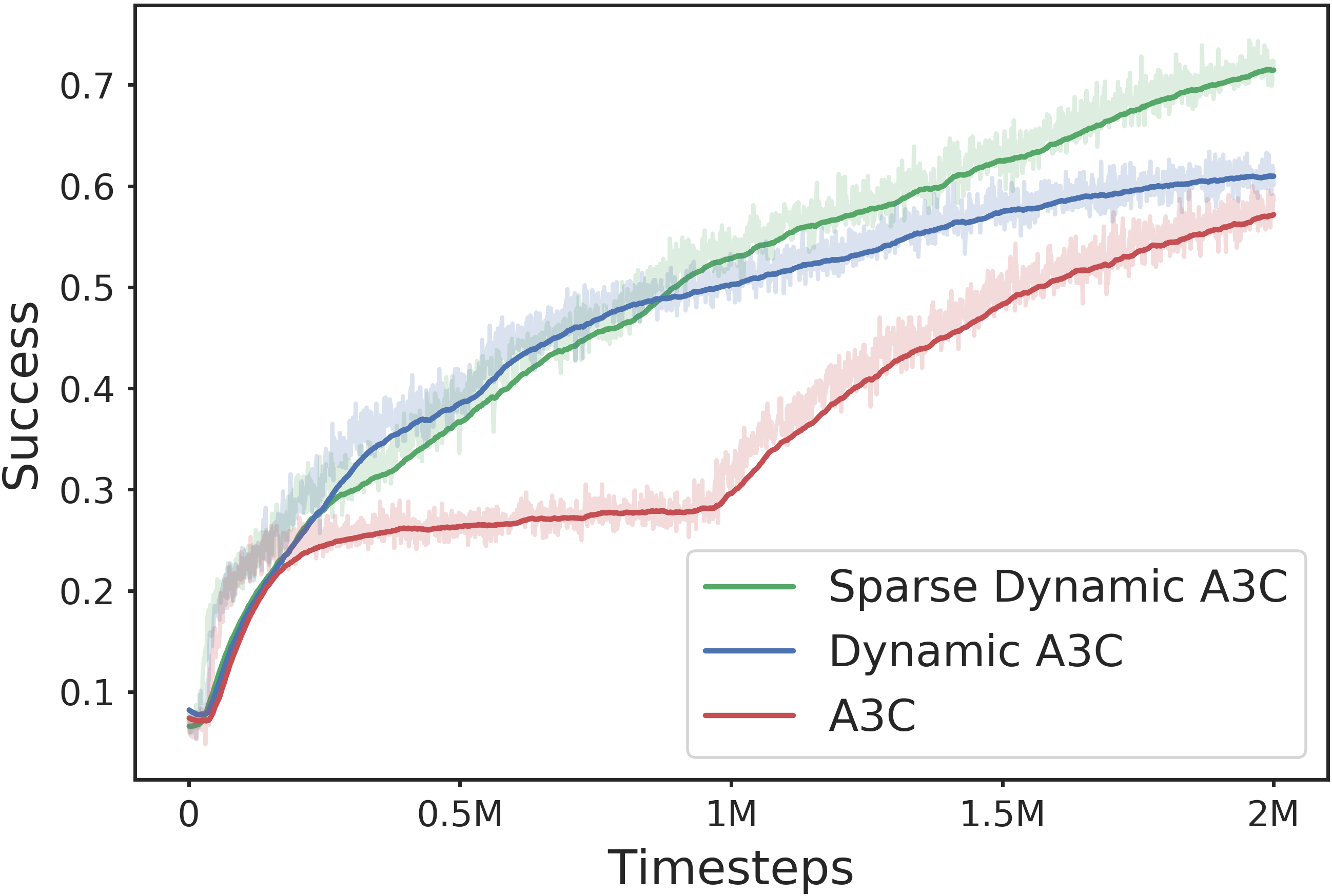}
		\vskip -0.1in
		\captionof{figure}{Comparing sample efficiency at training time for A3C and Dynamic A3C model on the visual navigation task.}
        \label{fig:visnav_train_curve}
	\end{minipage}
\end{table}



\section{Policy Gradient Formulation for Multi-Scene Environments}
\label{policy_grad}
In this section, we provide a mathematical derivation for extending the variance reduction formulation for policy gradient algorithms to multi-scene environments. As stated in the main paper, the derived / improved formulation shows that the sample variance for a multi-scene environment is best minimized by learning a joint value function $V(s,\mathcal{M})$ dependent on both state $s$ and MDP $\mathcal{M}$.

\subsection{Policy Gradient Equation}
Policy gradient algorithms aim to optimize the policy $\pi(a|s)$ by maximizing the expected trajectory reward. For a multiple-scene environment operating over an MDP set ${\mathcalbf{M}}: \{\mathcal{M}_1,\mathcal{M}_2, ... \mathcal{M}_N\}$, the expected reward is over different trajectories $\tau$ and the set of possible MDPs $\mathcalbf{M}$. Thus,
\begin{equation}
    J = \mathbf{E}_{\tau,\mathcal{M}}[\mathcal{R}_{\tau,\mathcal{M}}],
\end{equation}
where $\mathcal{R}_{\tau,\mathcal{M}}$ is the total reward for a trajectory $\tau$ sampled using MDP $\mathcal{M}$. The above equation can be broken down as,
\begin{align}
    &J = \sum_{\mathcal{M}} \sum_\tau P(\tau,\mathcal{M}) \ \mathcal{R}_{\tau,\mathcal{M}}. \\
    &J  = \sum_{\mathcal{M}} P(\mathcal{M}) \left(\sum_\tau P(\tau \vert \mathcal{M}) \ \mathcal{R}_{\tau,\mathcal{M}}\right).
\end{align}

The term $\sum_\tau P(\tau \vert \mathcal{M}) \ \mathcal{R}_{\tau,\mathcal{M}}$ can be understood as the expected reward over trajectories given a fixed underlying MDP $\mathcal{M}$. We denote this conditional expectation using $J_\mathcal{M}$. Thus,
\begin{align}
    J_\mathcal{M} = \sum_\tau P(\tau \vert \mathcal{M}) \ \mathcal{R}_{\tau,\mathcal{M}} \label{eq:single_conditonal}\\
    \implies J = \sum_{\mathcal{M}} P(\mathcal{M}) \ J_\mathcal{M}.
\end{align}

Computing gradients for both sides with respect to policy parameter $\theta$:
\begin{align}
    \nabla_\theta J =  \sum_{\mathcal{M}} P(\mathcal{M}) \ \nabla_\theta(J_\mathcal{M})\label{eq:policy_grad_div}.
\end{align}

Adopting single-MDP policy gradient derivation procedure from \cite{sutton2018reinforcement} for Eq. \ref{eq:single_conditonal}, the gradient with respect to $J_\mathcal{M}$ can be written as:
\begin{align}
    &\nabla_\theta(J_\mathcal{M}) = \sum_s \sum_a P(s,a|\mathcal{M})  \ [\nabla_\theta \log \pi(a|s)] \  Q(s,a,\mathcal{M}),\label{eq:single_policy_grad_extension}
\end{align}

where $\pi(a|s)$ represents the scene-generic policy function and $Q(s,a,\mathcal{M})$ is the action value function for the specific MDP $\mathcal{M}$.
Combining Eq. \ref{eq:policy_grad_div} and \ref{eq:single_policy_grad_extension},
\begin{align}
    &\nabla_\theta J = \sum_{\mathcal{M}} P(\mathcal{M}) \ \sum_s \sum_a P(s,a|\mathcal{M})  \ [\nabla_\theta \log \pi(a|s)] \  Q(s,a,\mathcal{M}). \\
    &\nabla_\theta J = \sum_{\mathcal{M}} \sum_s \sum_a P(s,a,\mathcal{M})  \ [\nabla_\theta \log \pi(a|s)] \  Q(s,a,\mathcal{M}). \label{eq:policy_grad_prob_sum}
\end{align}

Writing Eq. \ref{eq:policy_grad_prob_sum} in the form of an expectation mean, the expression for policy gradients in multi-scene environments is given by,
\begin{equation}
     \boxed{\nabla_\theta J = \mathbf{E}_{s,a,\mathcal{M}} \left[\left(\nabla_\theta \log \pi(a|s)\right) \  Q(s,a,\mathcal{M})\right]} \label{eq:policy_grad_mean}.
\end{equation}

\subsection{Variance Reduction}
The goal of variance reduction is to find a distribution with same mean as Eq. \ref{eq:policy_grad_mean}, but reduced variance so as to be able to compute the sample mean with minimum error using a limited number of samples. In this section, we derive the optimal formulation for variance reduction in \emph{multi-scene} environments.

\textbf{Lemma 1.} $Q(s,a,\mathcal{M})$ can be replaced by $\psi(s,a,\mathcal{M}) = Q(s,a,\mathcal{M}) - f(s,\mathcal{M})$, where $f$ is any general function, without affecting the overall policy gradient from Eq. \ref{eq:policy_grad_mean}.

\textbf{Proof:}
Let the new policy gradient be $\nabla_\theta J'$,
\begin{align}
     &\nabla_\theta J' = \mathbf{E}_{s,a,\mathcal{M}} \left[\left(\nabla_\theta \log \pi(a|s)\right) \  \psi(s,a,\mathcal{M})\right]. \\
     &\nabla_\theta J' = \mathbf{E}_{s,a,\mathcal{M}} \left[\left(\nabla_\theta \log \pi(a|s)\right) \  (Q(s,a,\mathcal{M})-f(s,\mathcal{M}))\right]. \\
     &\nabla_\theta J' = \nabla_\theta J - \mathbf{E}_{s,a,\mathcal{M}} \left[\left(\nabla_\theta \log \pi(a|s)\right) \  f(s,\mathcal{M})\right].\\
     &\nabla_\theta J' = \nabla_\theta J - \sum_\mathcal{M}\sum_s\sum_a P(s,a,\mathcal{M}) \left(\nabla_\theta \log \pi(a|s)\right) \  f(s,\mathcal{M}).\\
     &\nabla_\theta J' = \nabla_\theta J - \sum_\mathcal{M}\sum_s P(s,\mathcal{M}) \ f(s,\mathcal{M}) \ \sum_a P(a \vert s,\mathcal{M}) \ [\nabla_\theta \log \pi(a|s)].
\end{align}

Since we learn a common policy $\pi(a|s)$ for the entire MDP-set $\mathcalbf{M}$, $P(a \vert s,\mathcal{M}) = \pi(a|s)$,
\begin{align}
     &\nabla_\theta J' = \nabla_\theta J - \sum_\mathcal{M}\sum_s P(s,\mathcal{M}) \ f(s,\mathcal{M}) \ \sum_a \pi(a|s) \ \nabla_\theta \log \pi(a|s).\\
     &\nabla_\theta J' = \nabla_\theta J - \sum_\mathcal{M}\sum_s P(s,\mathcal{M}) \ f(s,\mathcal{M}) \ \left(\sum_a \ \nabla_\theta \ \pi(a|s)\right).\\
     &\nabla_\theta J' = \nabla_\theta J - \sum_\mathcal{M}\sum_s P(s,\mathcal{M}) \ f(s,\mathcal{M}) \ \left( \nabla_\theta \sum_a \ \pi(a|s)\right).\\
     &\nabla_\theta J' = \nabla_\theta J - \sum_\mathcal{M}\sum_s P(s,\mathcal{M}) \ f(s,\mathcal{M}) \ \cancelto{0}{\left( \nabla_\theta .1\right)}.\\
     &\implies \boxed{\nabla_\theta J' = \nabla_\theta J}.
\end{align}

Hence, without loss of generality we can write the multi-scene policy gradient equation as,
\begin{equation}
    \boxed{\nabla_\theta J = \mathbf{E}_{s,a,\mathcal{M}} \left[\left(\nabla_\theta \log \pi(a|s)\right) \  (Q(s,a,\mathcal{M})-f(s,\mathcal{M}))\right]}.
\end{equation}

\textbf{Lemma 2.} The optimal function $f(s,\mathcal{M})$ for minimizing the policy gradient sample variance is equal to the joint value function $V(s,\mathcal{M})$. 

\newpage
\textbf{Proof:} For a given tuple $ \{s,a,\mathcal{M}\}$, the sample value $\mathcal{X}$ is given by $\left[\nabla_\theta \log \pi(a|s)\right] \  \psi(s,a,\mathcal{M})$. Thus, the sample variance can be written down as,

\begin{align}
     &Var(\mathcal{X}) = E[\mathcal{X}^2] - (E[\mathcal{X}])^2.\\ \nonumber\\
     &\text{Given that the function $f(s,\mathcal{M})$ is parameterized by $\phi$:}\nonumber\\
     &\argmin_\phi Var(\mathcal{X}) = \argmin_\phi \left(E[\mathcal{X}^2] - (E[\mathcal{X}])^2\right).\\ \nonumber\\
     &\text{From \textbf{Lemma 1}, we know that the expected value of the samples remains unchanged for any function $f$.}\nonumber\\
     &\implies \argmin_\phi Var(\mathcal{X}) = \argmin_\phi E[\mathcal{X}^2],\\
     &\argmin_\phi Var(\mathcal{X}) = \argmin_\phi \mathbf{E}_{s,a,\mathcal{M}} \left[\left(\left(\nabla_\theta \log \pi(a|s)\right) \  \psi(s,a,\mathcal{M})\right)^2\right].\\ \nonumber\\
     &\text{Assuming independence between policy network gradients and the action value function,}\nonumber\\
     &\argmin_\phi Var(\mathcal{X}) = \argmin_\phi \mathbf{E}_{s,a,\mathcal{M}} \left[\left(\nabla_\theta \log \pi(a|s)\right)^2\right] \ \mathbf{E}_{s,a,\mathcal{M}} \left[\psi^2(s,a,\mathcal{M})\right],\\
     &\argmin_\phi Var(\mathcal{X}) = \argmin_\phi \mathbf{E}_{s,a,\mathcal{M}} \left[\psi^2(s,a,\mathcal{M})\right],\\
     &\argmin_\phi Var(\mathcal{X}) = \argmin_\phi \mathbf{E}_{s,a,\mathcal{M}} \left[\left(Q(s,a,\mathcal{M})-f_\phi(s,\mathcal{M})\right)^2\right].
     \\ \nonumber\\ \nonumber\\
     &\text{For optimal $\phi, \quad \nabla_\phi Var(\mathcal{X}) = 0$} \\
     & \implies \nabla_\phi Var(\mathcal{X}) \equiv  \nabla_\phi \ \mathbf{E}_{s,a,\mathcal{M}} \left[\left(Q(s,a,\mathcal{M})-f_\phi(s,\mathcal{M})\right)^2\right] = 0.\\
     & 2 \ \nabla_\phi f \ . \  \mathbf{E}_{s,a,\mathcal{M}} \left[f_\phi(s,\mathcal{M}) - Q(s,a,\mathcal{M})\right] = 0.\\
     &\sum_\mathcal{M} \sum_s \sum_a P(s,a,\mathcal{M}) \left[f_\phi(s,\mathcal{M}) - Q(s,a,\mathcal{M})\right] = 0. \\
     &\sum_\mathcal{M} \sum_s P(s,\mathcal{M}) \sum_a P(a \vert s,\mathcal{M}) \left[f_\phi(s,\mathcal{M}) - Q(s,a,\mathcal{M})\right] = 0. \\
     &\sum_\mathcal{M} \sum_s P(s,\mathcal{M})  \left[f_\phi(s,\mathcal{M}) - \sum_a \pi(a \vert s)\  Q(s,a,\mathcal{M})\right] = 0. \\
     &\sum_\mathcal{M} \sum_s P(s,\mathcal{M})  \left[f_\phi(s,\mathcal{M}) -  V(s,\mathcal{M})\right] = 0. \\
     &\mathbf{E}_{s,\mathcal{M}} \left[f_\phi(s,\mathcal{M}) -  V(s,\mathcal{M})\right] = 0.
    \\ \nonumber \\
     &\text{The above relation has to hold for any batch sample across $(s,\mathcal{M})$, which implies that,}\nonumber \\
     &\boxed{f_\phi(s,\mathcal{M}) = V(s,\mathcal{M})}.
\end{align}

\vskip 0.2in
Thus using \textbf{Lemma 1 \& 2}, we can write the optimal policy gradient formulation leading to minimum sample variance for a \emph{multi-scene} environment as,
\begin{equation}
    \boxed{\nabla_\theta J = \mathbf{E}_{s,a,\mathcal{M}} \left[\left(\nabla_\theta \log \pi(a|s)\right) \  (Q(s,a,\mathcal{M})-V(s,\mathcal{M}))\right]}. \label{eq:multi_policy_grad_mean}
\end{equation}

\section{Relationship between sample variance and value prediction error}
\textbf{Theorem 1.} The sample variance ($\nu$) for policy gradients defined by Eq.~\ref{eq:multi_policy_grad_mean}, can be minimized by reducing the prediction error $\epsilon$ between the true joint value function $V(s,\mathcal{M})$ and the predicted estimate $\hat{V}(s,\mathcal{M})$, where
\begin{equation}
    \epsilon = \mathbf{E}_{s,\mathcal{M}} \left[\left(V(s,\mathcal{M}) - \hat{V}(s,\mathcal{M})\right)^2\right].
\end{equation}

\textbf{Proof:}
The equivalent sample variance ($\nu$) for policy gradients defined by Eq.~\ref{eq:multi_policy_grad_mean}, can be approximated as,
\begin{align}
    \nu  &\approx \kappa \  . \ \mathbf{E}_{s,a,\mathcal{M}} \left[\psi^2(s,a,\mathcal{M})\right] \\
    &= \kappa  \ . \  \mathbf{E}_{s,a,\mathcal{M}} \left[\left(Q(s,a,\mathcal{M})-\hat{V}(s,\mathcal{M})\right)^2\right], 
    \label{eq:var1}
\end{align}

where $\kappa = \mathbf{E}_{s,a,\mathcal{M}} \left[\left(\nabla_{\theta} \log\pi(a|s)\right)^2\right]$ and $\hat{V}(s,\mathcal{M})$ represents the predicted value function. Now, using the true value function $V(s,\mathcal{M}) = \mathbf{E}_{a}\left[Q(s,a,\mathcal{M})\right]$, Eq. \ref{eq:var1} can be decomposed as,
\begin{equation}
    \nu  \approx \kappa  \ . \  \mathbf{E}_{s,a,\mathcal{M}} \left[\left(Q(s,a,\mathcal{M}) - V(s,\mathcal{M}) \  + \  V(s,\mathcal{M}) - \hat{V}(s,\mathcal{M})\right)^2\right]
\end{equation}
\begin{equation}
    \nu \approx \underbrace{\kappa  \ . \  \mathbf{E}_{s,a,\mathcal{M}} \left[\left(Q(s,a,\mathcal{M})-V(s,\mathcal{M})\right)^2\right]}_{\text{minimal possible variance}} +
    \underbrace{\kappa  \ . \ \mathbf{E}_{s,\mathcal{M}} \left[\left(V(s,\mathcal{M}) - \hat{V}(s,\mathcal{M})\right)^2\right]}_{\text{prediction error}} +
    \cancelto{0}{(\dots)}. \label{eq:var2}
\end{equation}
Thus, we see that the policy gradient sample variance $(\nu)$ can be minimized by reducing the error between the true value function $V(s,\mathcal{M})$ and the predicted estimate $\hat{V}(s,\mathcal{M})$.

\section{Clustering Hypothesis: Estimating MDP-Specific Value Function}

\subsection{True Value Function}
\label{mdp_specific_vf_est}
In order to make any comments about the distribution of the value function $V(s,\mathcal{M})$ over $s \in \mathcal{S} \ \& \  \mathcal{M} \in \mathcalbf{M}$, we first need to be able to compute the true scene-specific value estimates for any state $s \in \mathcal{S}$ and a particular MDP $\mathcal{M}$. While training an expert model for each level (as in \cite{parisotto2015actor}) is possible, it suffers with the following flaws,

\begin{enumerate}
\item The resulting value estimates $\{V_1(s),V_2(s), ... V_n(s)\}$ would not be over the same policy $\pi$.
\item The learned CNN backbone for each expert model will only be able to correctly process state images belonging to the corresponding level. This would lead to erroneous true value predictions for state images belonging to other levels.
\end{enumerate}

We instead propose the following strategy for estimation of the true MDP-specific value function,
\begin{itemize}
\item Train a general actor critic model over the entire MDP-set $\mathcalbf{M}$ to learn an intermediate but sufficiently good policy $\pi$. 
\item Freeze all the weights except the non-shared critic branch (usually a single fc layer)
\item Fine-tune this semi-frozen network, individually on each MDP $\mathcal{M}_i$ using only the critic value loss to get the true MDP-specific value function estimates.
\end{itemize}


\subsection{LSTM based Value Function}
\label{lstm_vf_est}
While demonstrating the insufficiency of traditional CNN/LSTM based networks, we argue that the value function learned by a CNN/LSTM network does not capture multiple dominant modes exhibited by the true value distribution. In this section, we outline the process used for computing the scene-specific value estimates learned by a CNN-LSTM critic network which was jointly trained on the entire MDP-set $\mathcalbf{M}$.
    
\begin{itemize}
    \item A CNN-LSTM network trained upto an intermediate policy $\pi$, is used to generate a dataset consisting of tuples of $\{$state images $s$, value estimates $\hat{V}(s)$ and CNN features $\phi(s)\}$. The associated samples are collected over 5k episode runs with the multi-level environment.
    \item In order to compute the value distribution samples for a target state $s_t$, we first extract a set of 200 state images $(\mathcal{S}_c: \{s_1,s_2 \dots s_{200}\})$, most similar to the target state in terms of the learned CNN features $\phi$. These states are then filtered to remove possible duplicates by setting a minimum threshold distance $||\phi(s_t) - \phi(s)||$ in the feature space $\phi$.
    \item As a final check, each state $s \in \mathcal{S}_c$ is manually examined to ensure semantic similarity with the target state $s_t$.
    \item Finally, the learned scene-specific value distribution is analysed by computing the histogram for the value estimates $\hat{V}(s)$ for $s \in \mathcal{S}_c$.
\end{itemize}

\section{Finding the Optimal Number of Dynamic Nodes}
\label{find_ncluster}
A critical component in training the dynamic model is the selection of number of dynamic nodes / clusters $N_b$. While it is possible to treat it as another hyper-parameter, we present an alternate approach for the same. This approach not only gives us the number of optimal clusters, but also strengthens our belief in the function of the proposed critic network.

Recall that, given our motivation from the clustering hypothesis, we believe that the true scene-specific value function distribution resembles a Gaussian Model Mixture (GMM). The final dynamic model needs to learn this distribution not just across different MDPs belonging to the set $\mathcalbf{M}$ but also for all states $s \in \mathcal{S}$. Thus, to find out the optimal number of gaussian clusters, we begin by approximating the multi-variate distribution $V(s,\mathcal{M})$ using discrete samples $\{s_j,\mathcal{M}_i,V(s_j,\mathcal{M}_i)\}$ for $s_j \in S \ \& \ \mathcal{M}_i \in \mathcalbf{M}$.

For the sample collection, we first obtain an intermediate policy $\pi$ and MDP specific value estimate networks $\hat{V}_i(s)$ for a random selection of 500 levels from the CoinRun ProcGen environment, using the strategy from Section \ref{mdp_specific_vf_est}. While incorporating the entire state space is infeasible, we try to minimize the error by sampling a large collection of 1000 states from the different levels using the common policy $\pi$. Next for each of these states $s_j, \ j \in [1,1000]$, we obtain the corresponding level specific value function estimates using $\hat{V}_i(s_j), i \in [1,500]$. 

\begin{figure}[H]
\floatbox[{\capbeside\thisfloatsetup{capbesideposition={right,center},capbesidewidth=5cm}}]{figure}[\FBwidth]
{\caption{\textbf{Finding optimal number of dynamic clusters.} AIC score curves for different number of clusters and training levels N. The local minima represents the Bayesian optimal choice \cite{forster2011aic} for the number of required dynamic / cluster nodes $N_b$.}\label{fig:ncluster_aic_coinrun}}
{\includegraphics[width=7cm]{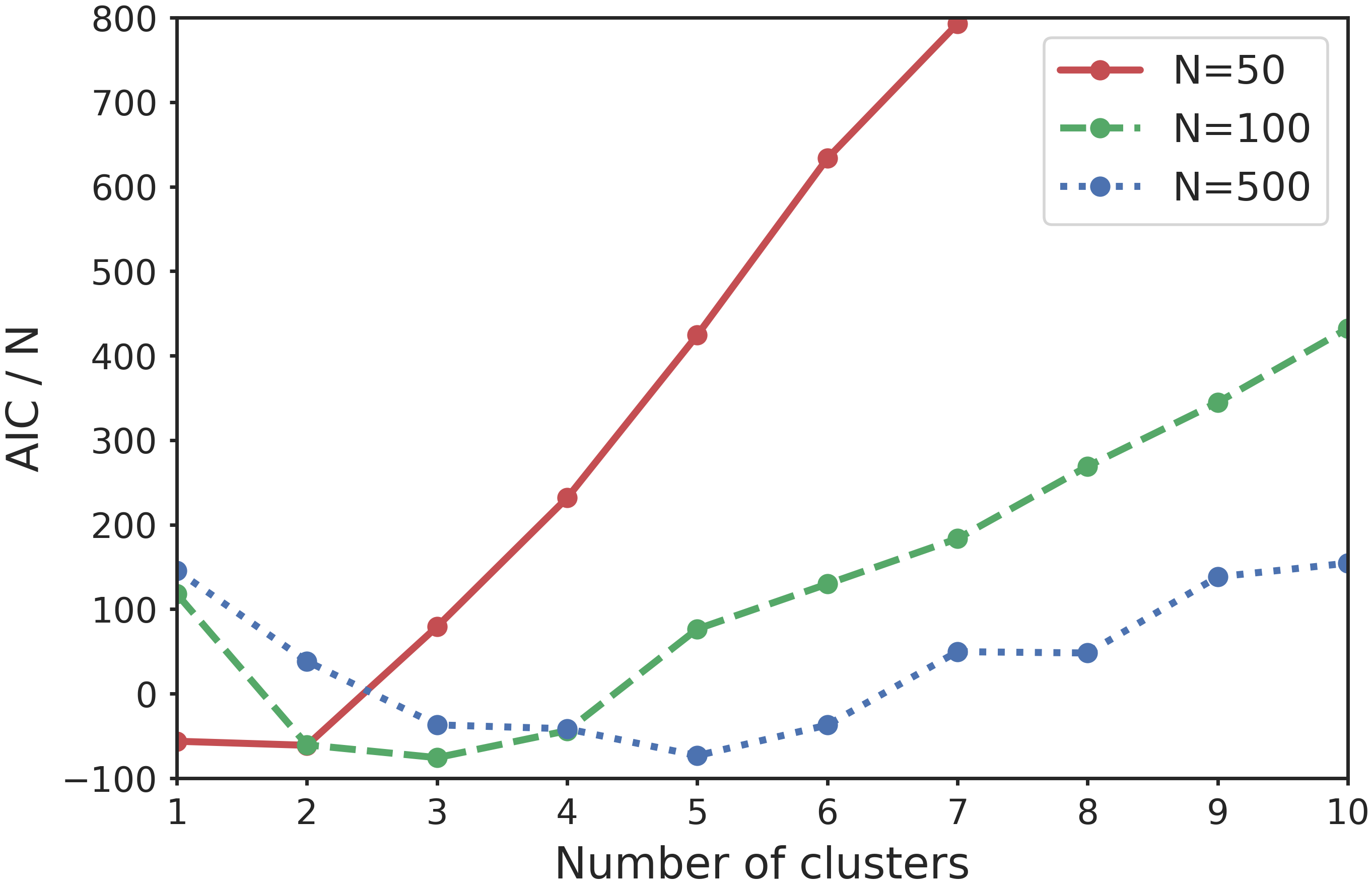}}
\end{figure}

The generated dataset representing samples from the multi-variate distribution $V(s,\mathcal{M})$ has shape [N $\times$ K], where N is number of levels and K is the number of states. The dataset is then fitted using a GMM with variable number of components $C \in [1,10]$. We finally use the Akaike Information Criterion (AIC) from \cite{akaike1987factor}, to determine the optimal choice for the number of clusters (refer Fig.~\ref{fig:ncluster_aic_coinrun}). 

\textbf{Result.} As shown in Fig.~\ref{fig:ncluster_aic_coinrun}, the optimal number of clusters (point of minima in the AIC/N curve) increases with the number of training levels. We also note that the AIC/N curve becomes less steep, as the number of training levels increases. This implies that given a sufficient number of training levels, the dynamic model's performance shows low sensitivity (higher robustness) to the selection of the hyper-parameter $N_b$. 

\end{appendices}
\end{document}